\documentclass[runningheads]{llncs}
\usepackage{graphicx}
\usepackage[utf8]{inputenc}
\usepackage{algorithm}
\usepackage{amsmath}
\usepackage[noend]{algpseudocode}
\usepackage{textcomp}

\makeatletter
\def\BState{\State\hskip-\ALG@thistlm}
\makeatother

\begin{document}

\title{An automatic text classification method based on hierarchical
taxonomies, neural networks and document embedding: the NETHIC tool.}

\author{Luigi Lomasto \inst{1,\star} \and
Rosario Di Florio\inst{2,\star} \and
Andrea Ciapetti\inst{3} \and
Giuseppe Miscione\inst{3} \and
Giulia Ruggiero\inst{3} \and
Daniele Toti\inst{3,4,\star\star}\orcidID{0000-0002-9668-6961}}

\institute{Eustema S.p.A. \email{l.lomasto@eustema.it} \and
Allianz SE \email{rosario.di-florio@allianz.com} \and
Innovation Engineering S.r.l., Rome, Italy \email{\{a.ciapetti,g.miscione,g.ruggiero,d.toti\}@innen.it} \and
Department of Sciences, Roma Tre University, Rome, Italy \email{toti@dia.uniroma3.it}}

\renewcommand{\thefootnote}{\fnsymbol{footnote}}
\footnotetext[1]{These two authors contributed equally to this work.}
\footnotetext[2]{To whom correspondence should be addressed.}

\maketitle

\begin{abstract}
This work describes an automatic text classification method implemented in a
software tool called NETHIC, which takes advantage of the inner capabilities of
highly-scalable neural networks combined with the expressiveness of hierarchical
taxonomies. As such, NETHIC succeeds in bringing about a mechanism for text
classification that proves to be significantly effective as well as efficient.
The tool had undergone an experimentation process against both a generic
and a domain-specific corpus, outputting promising results. On the basis of this
experimentation, NETHIC has been now further refined and extended by adding
a document embedding mechanism, which has shown improvements in terms of
performance on the individual networks and on the whole hierarchical model.
%
%
%
%

\end{abstract}

\section{Introduction} 
The last decade has seen an extremely high surge in the usage of network-based
technologies by people in their everyday lives, and as such an enormous amount
of information, a significant part of it in textual form, is being exchanged at a
constant rate.
As a matter of fact, social networks and online platforms are now an essential
way for people to share documents and data, but at the same time they are
leading to an increase of confusion and of potentially hidden or lost information.
In order to put some measure of order upon this deluge, methods and techniques
have arisen to try and provide users with means to classify the textual
information exchanged in a manner that may be as automatic as possible.
This is critical for an effective management and exploitation of the information
itself. Considerable efforts have been spent so far to solving this problem, by
bringing about corresponding solutions for text classification both in
literature and in commercial platforms.~\cite{autotextreview}.

In this regard, machine learning techniques such as supervised classification
can be effectively used to assign a number of predefined labels or classes to a
given textual document~\cite{Sebastiani}.

This work describes NETHIC, a software tool implementing an automatic text
classification method which, at its core, relies upon hierarchical taxonomies
and artificial neural networks (ANNs). The tool had been earlier introduced
in~\cite{NETHIC-ICEIS2019} in its original form, where its cross-domain
applicability had been shown by detailing an experimentation on both a general
purpose and a domain-specific classification task. The latter revolved around a
European funded project for detecting and analyzing criminal contents from
online sources.

In this paper, NETHIC's core elements and features are firstly reprised, and
then an extension to its methodology and functionality is described, which is
meant to improve the overall performance of the neural networks by means of a
Document Embedding mechanism, as proposed by Google.~\cite{Le-2014}.

Afterwards, a corresponding additional experimentation process is reported,
showing the expected improvements in terms of performance both on the individual
networks and on the whole hierarchical model.

%
The structure of this work is the following. 
In Section~\ref{section:Related Work}, related work is discussed. 
Section~\ref{section:Core elements of NETHIC} summarizes the core elements
making up NETHIC and provides a brief introduction to the Doc2Vec mechanism
introduced to enhance it.
Section~\ref{section:NETHIC's framework} describes the extensions of NETHIC's
original framework by detailing its current architecture, pre-processing, training and core algorithms used.
Section~\ref{section:Experimentation} reports a new experimentation showing a
comparison between NETHIC's original performance and the one resulting from the enhanced method.
Section~\ref{section:Conclusion} finally concludes the work and hints at future
developments.

\section{Related work}
\label{section:Related Work}
Classifying textual sources (documents) is a complex task that can be tackled by
computational methods like text mining and natural language processing.
These methods, as applied to different corpora and domains, include document
conceptualization and summarization~\cite{Toti-CONCEPTUM}, subject
categorization~\cite{Sebastiani}, sentiment analysis and author
recognition\cite{Wang,Koppel2}, and so forth.
The classical approach shared by a number of the aforementioned methods for
classifying texts is to represent them via high-dimensional vectors of features,
to be passed to machine-learning classifiers~\cite{Vidhya,Wang2}.
Vectors of features are built via a range of different techniques~\cite{Forman},
but the most common relies upon using frequencies of specific words or sets of
words (like n-grams, phrases, etc.) featured in documents within a corpus. These
frequencies can potentially be weighted as well. This technique is commonly
known as bag-of-words (BOW): typically, in such a technique keywords are derived
from training data, and a plethora of NLP methods is applied in order to do that,
including POS Tagging, Named Entity Recognition, Relation Detection and
others~\cite{Bird,Toti-PRAISED-BAMS,Toti-PRAISED-ICDE,Toti-PRAISED-EvoBio,Toti-PRAISED-ACM-BCB}.

In literature, different methods have been proposed to achieve optimal
classification performance. For example, Naive Bayes demonstrates the
effectiveness and efficiency for classifying test
documents~\cite{Lewis94acomparison,mccallum1998naive}, but it has poor performance when some
categories are sparse~\cite{Shen-2012}. As one of the deep learning algorithms,
recurrent neural network (RNN) is proposed by Pyo and Ha to deal with the
multi-class classification problem with unbalanced data~\cite{Ha-2016}, in which
the learnt word embedding depends on a recursive representation of the same initial feature space. In addition, convolutional neural network (CNN)
achieves remarkable performance in sentence-level
classification~\cite{kalchbrenner2014convolutional,kim2014convolutional,zhang-etal-2016-mgnc}.
Recently, CNN has been regarded as a replacement for logistic regression models
~\cite{zhang-wallace-2017-sensitivity}, which uses pre-trained word vectors as inputs for training the CNN
models~\cite{zhang-wallace-2017-sensitivity}. NETHIC therefore took this line as its preferred choice, in
order to achieve the best possible synthesis between correctness of results and
general performance of the classification system.
Hierarchical classification suffers from error propagation issue~\cite{Silla-2011}.
Zhihua Zhou et. al~\cite{Zhou:2006:TCN:1105850.1105888} show that only oversampling and threshold moving is
effective for training cost-sensitive neural networks by empirical studies. However, it
becomes difficult to define costs of misclassification when there are large
number of classes. A more recent paper~\cite{Buda-2017} also supports similar
claims, but suggests to use NLP and vector representation of sentences and words as a key,
to reduce the propagation of errors between categories in the hierarchy.
Choosing this approach, NETHIC and, in particular the extended version of the
tool, has chosen as a basis for the calculation of similarities a bag-of-word
approach and a multi-dimensional vector analysis of the structure of sentences
expressed in natural language. This allowed for a discreet but still significant
improvement over the previous version.

\section{Core elements of NETHIC} 
\label{section:Core elements of NETHIC}
This section summarizes NETHIC's core elements, by providing a brief description
of NETHIC's main approach, and by detailing the structure of the hierarchical
taxonomy needed for the tool to work and the datasets used in the training
phase.

\subsection{NETHIC's approach}
Taxonomies are data modeling structures able to describe knowledge in a way that
is both machine-processable and human-friendly. Their inner hierarchy can easily
serve the purpose of supporting the process of classifying contents either for a
human user or for a computational mechanism.~\cite{taxonomy1}.
Besides, ANNs are a class of machine-learning methodologies that has proven to
be significantly useful for discovering patterns among resources.

A combination of ANNs and taxonomies can therefore be extremely effective when
dealing with huge amounts of data, and can also scale pretty well on
multi-processor or multi-core hardware architectures, outperforming in terms of
processing time several other types of mechanisms sharing comparable levels of
effectiveness.~\cite{hermustad}.

This is exactly the approach used by NETHIC and the objective it tried to
achieve.

The earlier validation of the soundness and effectiveness of NETHIC's method had
been carried out on a corpus made up of Wikipedia articles representing subject
categories, including 500 articles for each category~\cite{NETHIC-ICEIS2019}.
Mechanisms used for the tokenization process include NLTK (Natural Language Processing
Toolkit) algorithms, whereas for extracting features, instead of encoding the
frequency of keywords, sentences have been decomposed into words, and the
latter have been turned into sequences of vectors and then passed to the deep learning
methods.
In this regard, a certain similarity is shared between NETHIC's approach and
other models based on probabilities, including Markov models, conditional random
fields and n-grams.

\begin{figure}[htbp]
	\centering
	\includegraphics[scale=1]{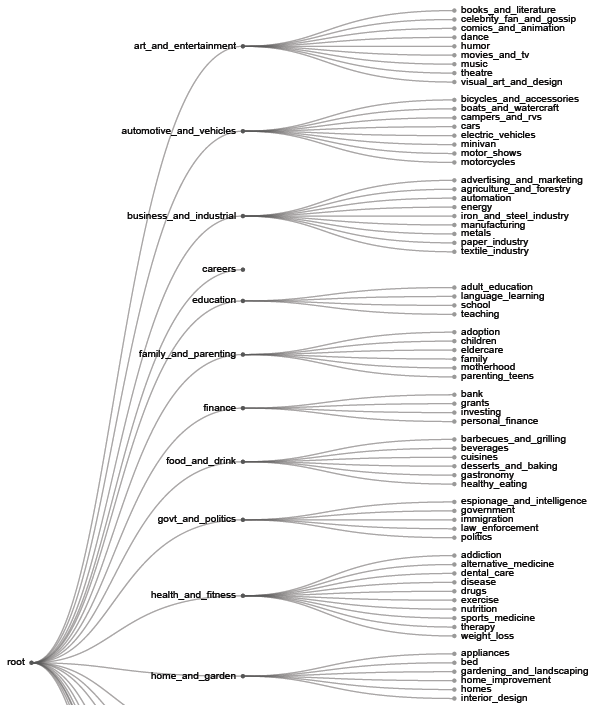}
	\caption{General-purpose hierarchical taxonomy (upper part).}
	\label{figure:General-purpose hierarchical taxonomy (upper part)}
\end{figure}

\begin{figure}[htbp]
	\centering
	\includegraphics[scale=1]{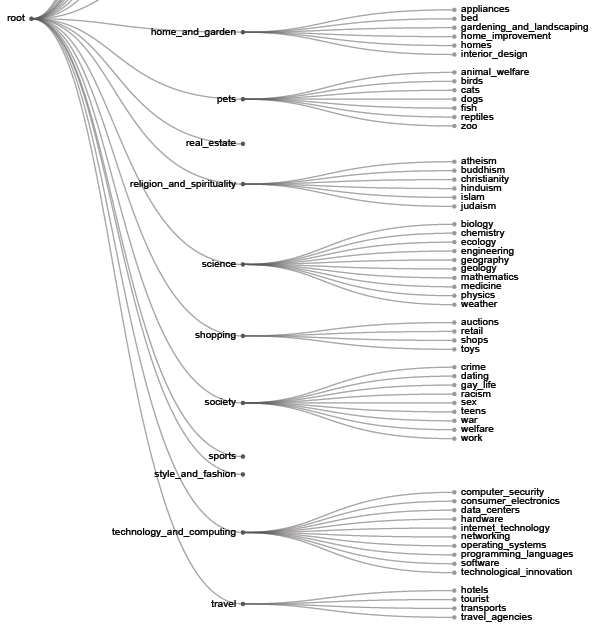}
	\caption{General-purpose hierarchical taxonomy (lower part).}
	\label{figure:General-purpose hierarchical taxonomy (lower part)}
\end{figure}

\subsection{Underlying Data Model}
\label{subsection:Hierarchical taxonomy} 	
As described in~\cite{NETHIC-ICEIS2019}, NETHIC's starting point lies in the
use of taxonomies, since their very nature as a hierarchical representation
contributes to place a certain level of order on top of unstructured or
semi-structured textual data. As such, with their clearly-defined logical
categories, they also make accessing and browsing such data dramatically
easier both for users and for software systems.

As a matter of fact, taxonomies play an essential role within NETHIC because
they are used as a bridge meant to connect its underlying knowledge model (with
data taken from Wikipedia) to subsequently build the datasets used during the
training phase of the method. These tree-like structures were shown to provide
the output and input classes for the respective input and output layer of each
neural network used in NETHIC, demonstrating their critical role in the training
phase.~\cite{NETHIC-ICEIS2019}.

While the research described in~\cite{NETHIC-ICEIS2019} reported the use of two
taxonomies, with one of them having the purpose of tackling a domain-specific
classification task, this work now focuses only on a general-purpose
taxonomy that covers a wide range of general topics,
whose upper and lower halves are displayed in Figure~\ref{figure:General-purpose
hierarchical taxonomy (upper part)} and Figure~\ref{figure:General-purpose
hierarchical taxonomy (lower part)}, respectively.

This taxonomy is defined in the RDF standard ~\cite{RDF}, starting from an
abstract classification class, here referred to as \textit{root}, and moving to
the lower class, here referred to as \textit{leaf}. This classification
structure contains 21 root child categories with a tree depth of 2, with a total of 117 leaf categories.
Not all the categories feature an expansion of the tree structure, {\it
i.e} not every sub-tree of the taxonomy possesses the same structure in
terms of the sub-categories of the classification.
\newline Each leaf is then manually connected to the associated category of
the Wikipedia graph by using the SKOS properties
\textbf{\texttt{skos:exactMatch}} and
\textbf{\texttt{skos:majorMatch}}~\cite{SKOS}, which support the subsequent
construction of the dataset explained in \ref{subsection:Dataset construction}.

\subsection{Dataset}
\label{subsection:Dataset construction}
NETHIC currently makes use of a dataset of 57,304 text documents, taken from
Wikipedia.
Wikipedia was chosen for this research because it provides an extensive
general-purpose knowledge archive that is regularly updated, as well as
being easily accessible on the Internet via its HTTP APIs.
As mentioned earlier, the taxonomy defined as NETHIC's core knowledge model is
used to provide the connections between the various classes and the Wikipedia
knowledge graph.
The first step of this process lies in downloading the entire library of
categories from Wikipedia and storing them in a graph structure.
These categories are used to group pages by related topics, and are used mostly
to find and navigate articles related to a particular
subject~\footnote{https://www.wikidata.org/wiki/Q2945159}~\footnote{https://en.wikipedia.org/wiki/Help:Category}.

Starting from the \textbf{\texttt{Category:Main\_topic\_classifications}}
category, the list of sub-categories and documents belonging to that
category is recursively retrieved by using the available APIs.
This first step results in a graph containing 1.5 million nodes linked together
with a subclassing relationship.

The next step of this process involves the computation of the feature vectors,
by using the category and sub-category names and, when available, their short
description, for each of the nodes in the graph via a word embedding approach.
Nodes and vectors are cached locally and the edges are weighted as
follows.

Given an edge \(e(u,v)\) and the respective vectors \(V_u\) and \(V_v\), the
weight \(w_e\) is:

\[w_e = inverse\_cosine\_similarity(V_u,V_v)\]

This allows NETHIC to have a weighted graph based on the semantic value of
Wikipedia's categories, here referred to as \textit{Wikipedia Category
Graph}.
The subsequent step of the process starts from the \textit{leaf} categories of
NETHIC's taxonomy and proceeds by navigating and collecting documents from
Wikipedia's categories following the shortest semantic path, until the desired amount of documents are collected.
The collected documents are then stored in a structure of folders and
sub-folders that follows the structure of NETHIC's taxonomy, where the
intermediate folders are formed by using a balanced amount of documents coming
from each \textit{leaf} category folder.

\subsection{Document embedding}
In the latest years, very important results has been achieved regarding text
representation to solve many NLP problems. One of the
most renowned solution is a word embedding model, called Word2Vec and proposed
by Google~\cite{Mikolov-2013}. A direct update of this model performs the
embedding of sentences or entire documents, and is known as Doc2Vec
~\cite{Le-2014}, which is useful to transform sentences or documents into
corresponding n-dimensional vectors. This transformation is pivotal
since it provides the possibility to work with documents without facing
the high dimensionality problem commonly present when a bag-of-words approach is
used for text representation. Another advantage for this type of word and
document representation lies in the semantic similarity as explained by the
authors in their studies. A generic application consists of comparing similar
words or document vectors through a cosine similarity metric, in order to
evaluate how close two items are in the semantic space.
NETHIC uses a Doc2Vec model, trained with an English Wikipedia corpus,
following two different strategies. In the first strategy, BOW
features are replaced with Doc2Vec vectors, used as features for
NETHIC's corpus, in order to verify if sufficiently good results could be
obtained with less information.
In the second strategy, the BOW functionality is merged with Doc2Vec, so
that it may be possible to use the occurrences of words in conjunction with the
semantic meaning of the documents.

\section{NETHIC's methodological and technological framework}
\label{section:NETHIC's framework}
This section reprises and extends NETHIC's framework with respect to the
earlier discussion in~\cite{NETHIC-ICEIS2019} in terms of its architecture,
pre-processing and training mechanisms and algorithms used, underlining the
improvements over its earlier version.

\subsection{Architecture}
\begin{figure}
	\centering
	\includegraphics[scale=0.5]{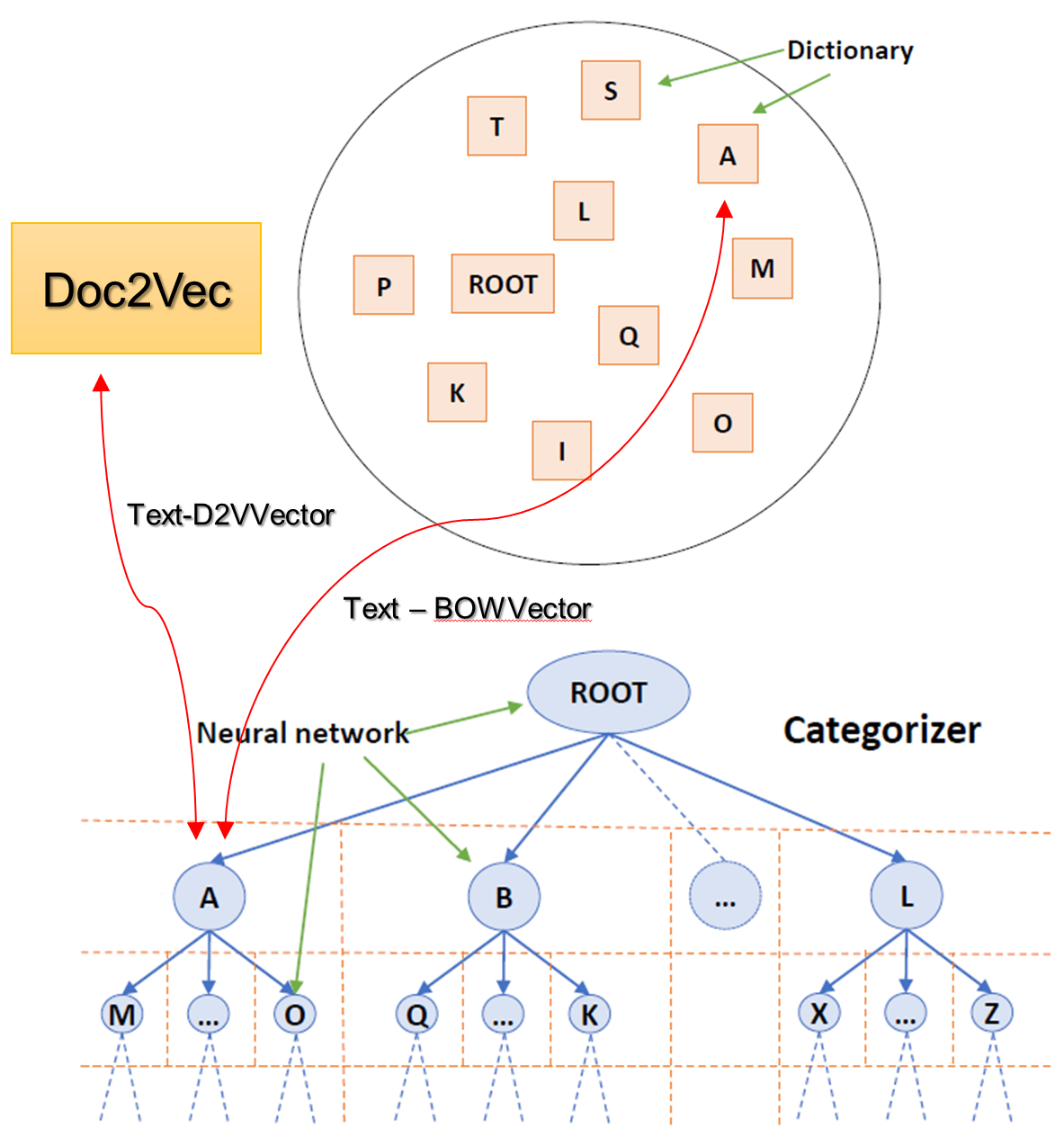}
	\caption{NETHIC's Architecture}
	\label{figure:architecture}
\end{figure}

The main components of NETHIC's current architecture are artificial neural
networks, a hierarchical taxonomy, dictionaries and a Doc2Vec pre-trained model.
Figure~\ref{figure:architecture} shows these components and graphically
represents the structure and NETHIC's whole process. The latter starts
with a text elaboration, by using dictionaries and a document embedding instance
to vectorize the input documents, and goes on by relying upon a hierarchical
neural networks model to find the main leaf categories to be used as classification
labels for the given documents.

\subsection{Reasons for a hierarchical neural network model}
In NETHIC, as initially clarified in~\cite{NETHIC-ICEIS2019}, a neural network
hierarchy is employed for several reasons. In fact, in order to classify a
document whose main topic is, for instance, \textit{kitchens}, it is sensible to
use a neural network that is trained only on texts whose focus is on interior
decoration and house supplies, instead of a more heterogeneous or too general
artificial neural network.

This avoids the presence of unnecessary words and reduces the noise on the
classification process. In NETHIC, for each taxonomy
concept, with the exception of the leaf concepts, one neural network is trained
and a dictionary of words is built. In the upper levels, the neural
networks trained are characterized by a somewhat horizontal view, splitting
documents according to general, wide concepts like Economy, Religion, Science
and Sports, whereas in the deeper levels the networks tend to assume a
more vertical separation and classify the documents according to a more specific
category that is a descendant of the general concept (for example, Sports),
{\it e.g. Basketball, Combat Sports, Golf, Soccer, Swimming, Tennis,
Volleyball...}

Thus, the classification function used in the neural networks, located at
different levels in the hierarchy, is trained with a vocabulary and a set of
words, having a varied logical structure and granularity. The upper
levels are trained with generic words, which are alike to general ``concepts'',
and the vocabulary used does not include the complete glossary of words
associated with the context. When reaching the deeper levels, instead, there is
a progressively extensive need to discriminate between semantically-close
concepts.

Therefore, the glossary used in the training process contains more
specific words, since the classification process, in order to be as effective as
possible, has the need to learn additional knowledge on the given area of
interest. That is why an iterative approach is followed, particularly suited
to artificial neural network-based methods, by descending to the more
specific, deeper levels of the classification.

This allows NETHIC to prevent the occurrence of semantic errors when dealing
with words belonging to different conceptual areas (like the word
{\it tree} (that represents a plant in the natural world, a component of a ship
or a data-representation structure in computer engineering).
This sometimes forces the process, when it tries to classify more generic
documents, to stop the iteration earlier before it reaches the deepest levels.

In order to understand the potential offered by this approach, let us consider
another example where a given document talks about \textit{advertising and
marketing}. By taking a look at the taxonomy shown
in~\ref{figure:General-purpose hierarchical taxonomy (upper part)} and~\ref{figure:General-purpose hierarchical
taxonomy (lower part)}, there are many concepts semantically close to the
given category such as \textit{personal finance}, \textit{shops} and
\textit{movies and tv}, each belonging to different paths. In a scenario where a
single neural network is used on 117 different classes, it can be easy to get
irrelevant results and low scores due to using the same words in different
contexts and with different meanings. In order to face these issues, a
hierarchical approach comes thus in handy to decompose the main problem into
many sub-classification problems, all of them working together to reduce the
noise due to the context by considering trained neural networks on semantically
distant concepts.

\subsection{Classification process}
The classification process starts with an unstructured
text/document as input. 

Initially, the \textit{root} category's dictionary and a Doc2Vec
(D2V) pre-trained model is used to transform text into a corresponding
vectorized form.
After that, a BOW+Doc2Vec composed vector is passed as input to \textit{root's
Neural Network} to perform the prediction task. As explained
in~\ref{subsection:Algorithm to build paths}, the first relevant categories are
chosen to continue with the next steps, considering appropriate dictionaries and
neural networks.

\subsection{Data Pre-processing}
The pre-processing step is required to transform the unstructured datasets
explained in \ref{subsection:Dataset construction} in order to obtain a useful
and structured version of the data. Before delving deeper into the
pre-processing step, it is noteworthy to say that the initial corpus has been
split into two balanced corpora with a ratio of 95\% - 5\%. The first corpus,
named \textit{Corpus-A} and containing 54.439 documents (about 465 for each
leaf category), has been used for the training and validation tasks on single
neural networks. The second corpus, called \textit{Corpus-B}, containing 2843
documents (about 25 for each leaf category), has been used throughout the entire
validation of the hierarchical model.
As known in literature as well as in commercial environments, an ETL
(Extraction, Transformation and Loading) process represents a key point for data collection
and feature extraction tasks. In this work, three kinds of transformations,
explained below, are used in order to build a sufficient number of datasets to
check and identify the best features to be used. 

The details of the first
transformation can be found in~\cite{NETHIC-ICEIS2019}; it produces
BOW-based datasets that will be referred to as \textit{Datasets\_BOW} from now
on. Dictionaries used in the hierarchical validation step are saved in
order to transform the validation corpus by considering the same words already
used to train the neural networks.

A second transformation used the Doc2Vec model to convert documents into
suitable vectors of 300 dimensions. Unlike the first transformation, the built
datasets called \textit{Datasets\_D2V} consume a really slight portion of memory
and there is no need to store dictionaries for the subsequent validation phase.

Finally, the two abovementioned transformations in order to use both
features type. In this case the datasets obtained called
\textit{Datasets\_D2V-BOW} are far too large to be kept in memory, just like the
BOW-based datasets. The resulting vectors will be in this case the concatenation
of the BOW and D2V vectors, thus dictionaries are saved here as well.

For each of these transformations, 18 datasets are built. 

The next subsection describes how three corresponding models of neural networks,
each for one of the three transformations, have been trained and compared.

\subsection{Training} 
As mentioned in the previous subsection, the training phase carried out by using
\textit{Corpus-A} has been performed three times, one for each transformation
(and thus for each group of datasets).
In this subsection, firstly the generic method used to train the single neural
networks will be explained, and then the different models will be compared in
order to find the best features to be used for the classification problem.

According to the best practices for training artificial intelligence models, 
a cross-validation was executed to check for potential
overfitting/underfitting, by using the k-fold and ``leave one out
approach''~\cite{Wong-2015}.
By resorting to this technique, an initial, balanced splitting of the datasets
has been necessary to compute the training and testing in ``leave one out''.
Starting from this assumption, for any single dataset two sub-datasets have been
built, with 90\% and 10\% proportions, respectively.
For example, considering a theoretical category \textit{X} and a corresponding
BOW dataset saved as \textit{Dataset\_BOW\_X}, a splitting is made in order to
obtain \textit{Dataset\_BOW\_X\_Training\_CV} and
\textit{Dataset\_BOW\_X\_Validation\_LOO}. The following pseudocode describes
how the training phase was performed.

\begin{algorithm}
	\caption{Training using Cross-Validation and Test in One Shot}\label{euclid}
	\label{alg:algorithm1}
	\begin{algorithmic}[1]
		\Procedure{Training}{}
			\For {\textit{middle Taxonomy's Category  \textbf{X}}}
				\State $\textit{Dataset\_X\_Training\_CV} \gets $\textit{Dataset\_X}
				\State $\textit{Dataset\_X\_Validation\_LOO} \gets $\textit{Dataset\_X}
				\State $\textit{CV\_accuracy} \gets $\textit{0}
				\For {\textit{each folds combinations (4,1) from
					\textit{Dataset\_X\_Training\_CV}}} \State $\textit{current\_CV\_model\_X}
					\gets $\text{training(4\_ folds)} \State $\textit{current\_CV\_accuracy\_X}
					\gets $\text{model.validation(1\_fold)} \State $\textit{CV\_accuracy} \gets
					$\textit{CV\_accuracy + current\_CV\_accuracy\_X}
				\EndFor
				\State $\textit{CV\_accuracy} \gets $\textit{CV\_accuracy : 5}
				\State $\textit{model\_X} \gets
				$\text{training(\textit{Dataset\_X\_Training\_CV})} \State
				$\textit{model\_accuracy\_X} \gets
				$\text{model.validation(\textit{Dataset\_X\_Validation\_LOO)}} \State
				\textbf{save(model\_X)} \emph{}
			\EndFor
		\EndProcedure
	\end{algorithmic}
\end{algorithm}

\begin{figure}
	\centering
	\includegraphics[scale=0.7]{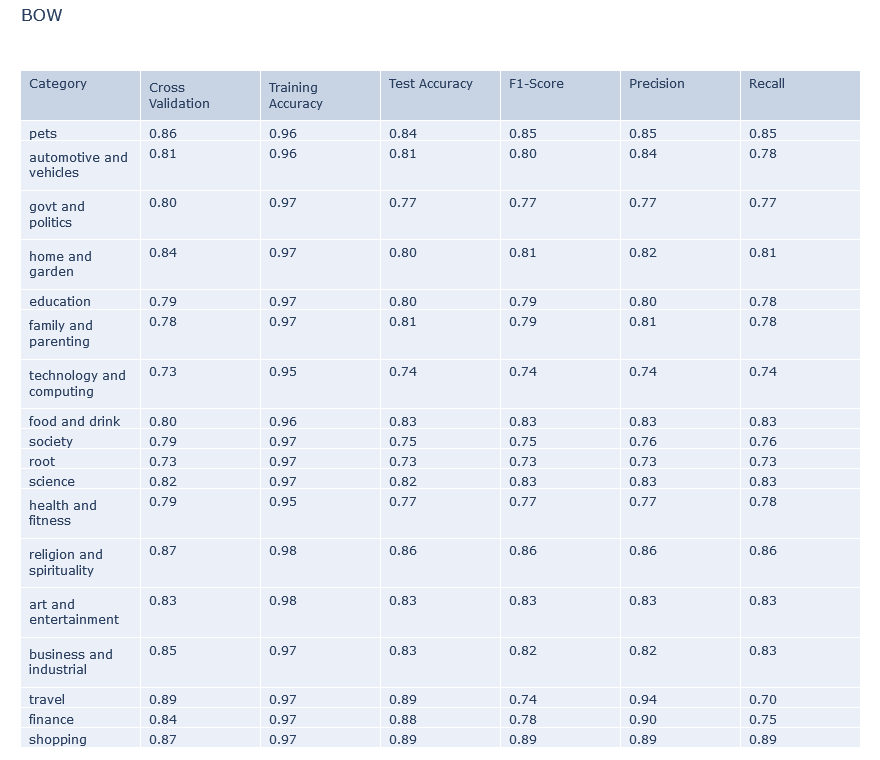}
	\caption{Single ANN's scores with BOW Dataset}
	\label{figure:bow_single}
\end{figure}

\begin{figure}
	\centering
	\includegraphics[scale=0.7]{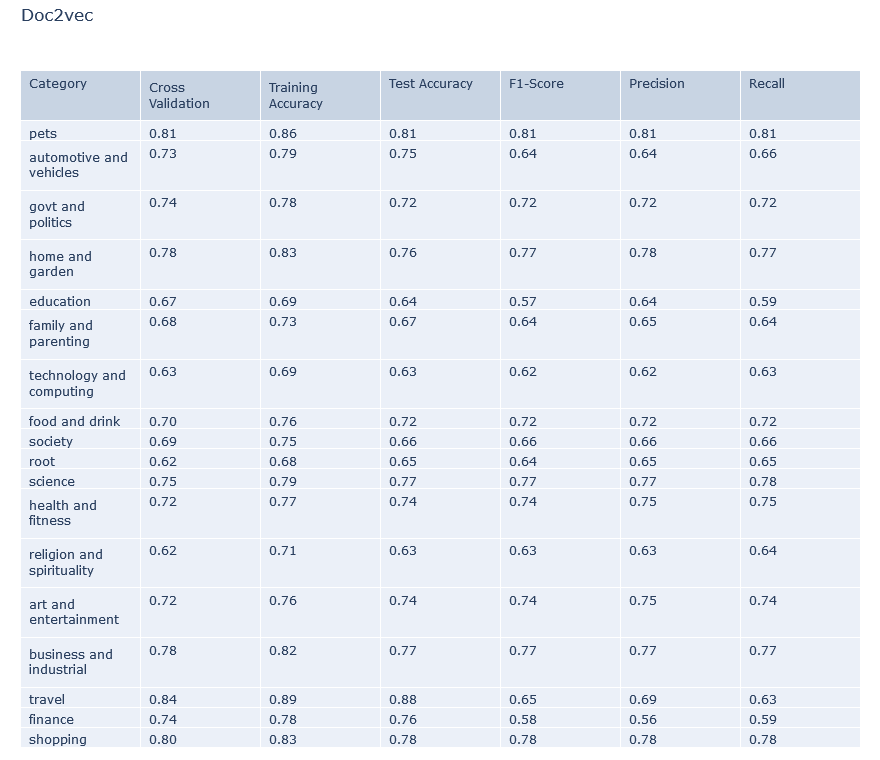}
	\caption{Single ANN's scores with Doc2Vec Dataset}
	\label{figure:doc2vec_single}
\end{figure}

\begin{figure}
	\centering
	\includegraphics[scale=0.7]{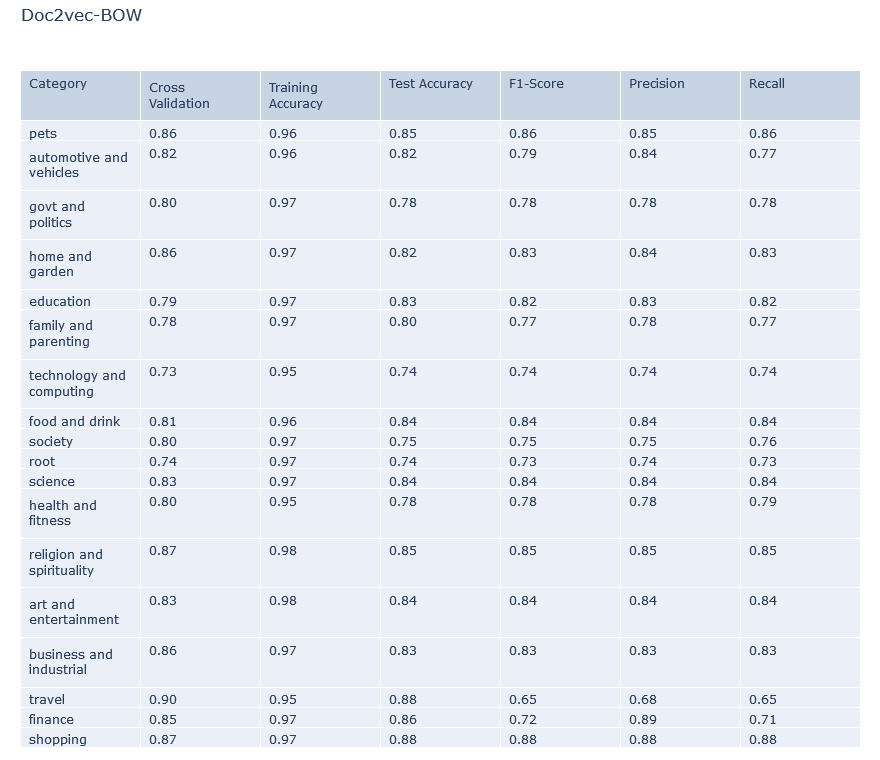}	
	\caption{Single ANN's scores with BOW+Doc2Vec Dataset}
	\label{figure:bow-doc2vec_single}
\end{figure}

Basically, any category used to realize the hierarchical model covers all
the steps described in~\ref{alg:algorithm1}, and for each of them, a
cross-validation has been performed in order to evaluate the
potential presence of underfitting and overfitting. After making sure that none
of these problems had arisen, it was possible to train the neural network using
\textit{Dataset\_X\_Training\_CV} subsequently validated with
\textit{Dataset\_X\_Validation\_LOO}. This algorithm has been executed on the
three groups of datasets previously described, obtaining neural networks for
each of the features considered, that is to say BOW, Doc2Vec and BOW-Doc2Vec.
The three tables showed in Figure~\ref{figure:bow_single},
Fig~\ref{figure:doc2vec_single} and Fig~\ref{figure:bow-doc2vec_single},
respectively, contain Cross-Validation Accuracy, Training Accuracy, Test
Accuracy, Precision, Recall and F1-score metrics for each of the trained models.
As shown, the best accuracies are obtained with the combined model that uses
BOW and D2V features together.  The worst performance was obtained by using
the model trained on D2V features only: this means that for this kind of
complex classification, document embedding by itself is not a good choice to
represent documents, but it can nevertheless be useful to improve the accuracy
of the BOW model, as seen in the results obtained. By considering the BOW and
BOW-D2V accuracy values, there is an improvement of about 2\% for most
categories, and an improvement of about 1\% for the \textit{root} category: this
is especially important, because in the hierarchical model it represents the
heaviest category for the correct construction of the classification paths.
Cross-validation results show that there are no overfitting and underfitting
issues exactly as expected. Training and test accuracies show that all
the trained models learn well and are able to generalize with data never seen
before. Precision, Recall and F1-Score show that the trained models are able to
obtain a good accuracy for all of the labels, and in general they do not
confuse among classes that are semantically close to one another.

\subsection{Algorithm to build paths}
\label{subsection:Algorithm to build paths}
The algorithm used to build paths has not undergone significant modifications
from the one described in~\cite{NETHIC-ICEIS2019}. For each returned path, the
average between all the single scores for each of the corresponding categories
is computed.

For instance, given the following path: \texttt{P}=
\texttt{C\textsubscript{1}}/\texttt{C\textsubscript{2}}/C\texttt{\textsubscript{3}}
with its respective scores \texttt{SC\textsubscript{1}},
\texttt{SC\textsubscript{2}} and \texttt{SC\textsubscript{2}}, its corresponding
total score \texttt{S\textsubscript{P}} is the average of the single scores.
The system keeps considering categories until the probabilities returned by the
current neural networks reach a threshold that is initially set as \texttt{0.7}.
If after the first classification a good current tolerance is obtained, this
will consequently lead to a reasonable classification; otherwise, if such a value is
low, it means that there are paths with a low score. In this case, a second
classification iteration is run by considering the paths with a lower score
value. In general, this algorithm allows the system to select the highest-level
categories and concepts when the textual content examined contains only generic
terms, whereas it is possible to select more detailed and low-level categories
and concepts by examining texts that are very specific, technical or focused on
a certain topic.

\section{Experimentation of NETHIC's extended method and comparison with the
earlier method}
\label{section:Experimentation}
In this section the results of the new experimentation carried out after the
introduction of the combined BOW+Doc2Vec document embedding mechanism is
reported and compared with the earlier version of NETHIC (with only the BOW
mechanism). The focus here is on the pie charts and confusion
matrices that show how integrating the Doc2Vec model for feature extraction is a
sound approach combined with the earlier BOW-based method. For the purposes of
such a comparison, the terms ``NETHIC'' and ``NETHIC-2'' will be used to differentiate
between NETHIC's original approach and the extended one, respectively.
The last part of this section discusses a couple of practical examples to
conclude the analysis.

\subsection{Comparison between NETHIC and NETHIC-2}
As explained in~\cite{NETHIC-ICEIS2019}, to evaluate the tool's accuracy
the first three categories returned by the algorithm to
build paths detailed in~\ref{subsection:Algorithm to build paths} are
considered.
This choice is meaningful because when many classes ---some of them semantically
close to one another --- are used for text classification, a single assigned
class may not be the only and optimal solution.
For this comparison \textit{Corpus\_B}, which contains 2843 documents (about 25
for each leaf category), has been used to test both methods. 

\begin{figure}
	\centering
	\includegraphics[scale=0.9]{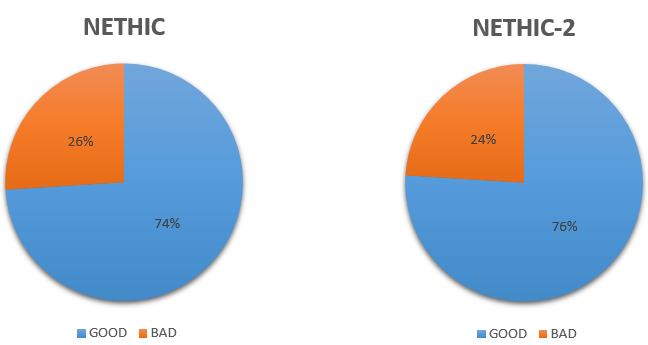}	
	\caption{Classification accuracy of the initial (leftmost chart) and the
	extended method (rightmost chart).}
	\label{figure:Models Accuracy}
\end{figure}

Clearly, dictionaries are used step-by-step for each different path in order to
build the correct BOW vector to be merged with the unchanged Doc2Vec vector
(which stays the same for every document to be classified), in order to keep
the coherence with the currently analyzed category.
The pie charts in Figure~\ref{figure:Models Accuracy} emphasize the
improvement obtained with the extended method, which is able to correctly
classify \texttildelow 60 documents more than the earlier approach.
The improvement observed during the training phase is the same as in this
evaluation, and therefore confirms an overall improvement of 2\%.

\begin{figure}[ht]
	\includegraphics[scale=0.25]{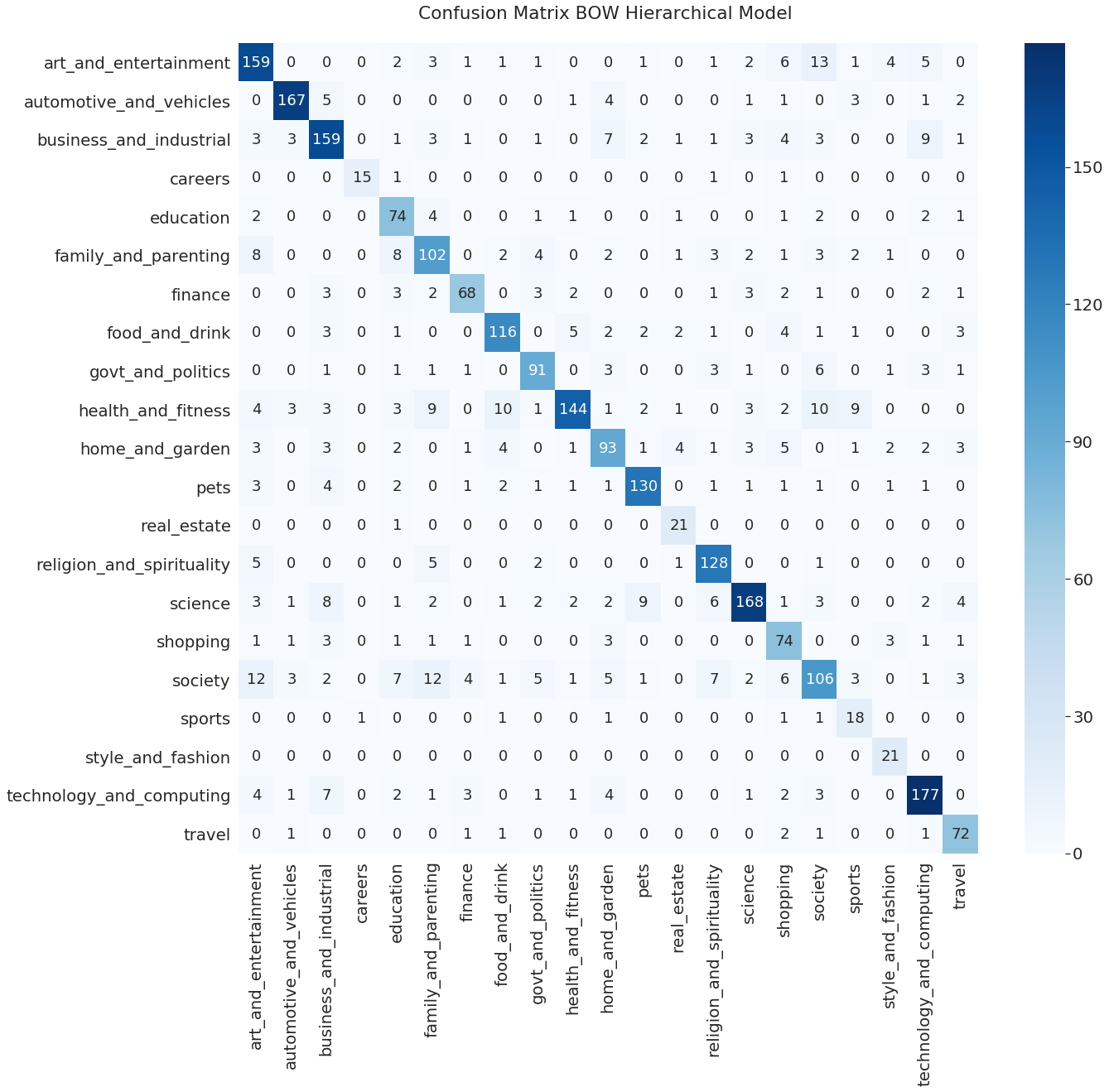}
	\caption{NETHIC's original results (with only the BOW-based embedding
	mechanism).}
	\label{figure:bow_hierarchical}
\end{figure}

\begin{figure}[htb]
	\includegraphics[scale=0.25]{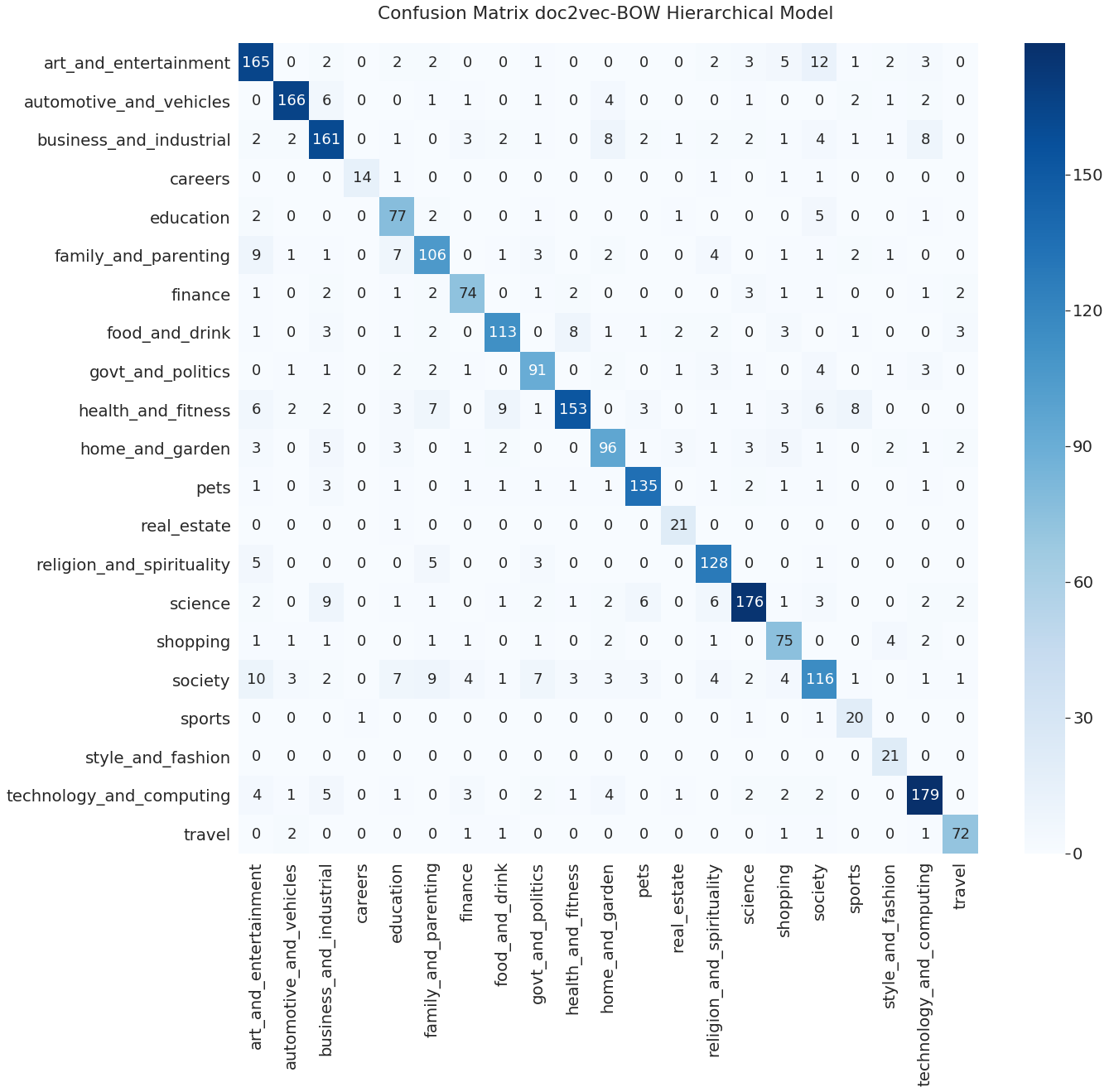}
	\caption{NETHIC's results with the introduction of the combined BOW+Doc2Vec
	embedding mechanism.}
	\label{figure:bow-doc2vec_hierarchical}
\end{figure}

The following confusion matrix shows the methods' accuracy for the first
hierarchical level in order to understand the improvement for the root neural
network.
The diagonal values for the matrix in both Figure
\ref{figure:bow_hierarchical} and \ref{figure:bow-doc2vec_hierarchical}
represents the correct classifications and make the matrix almost diagonal.
The best performance for the \textit{Science} category is obtained by
NETHIC-2 with about 8 more documents that with the earlier method had been lost. In general,
improvements over NETHIC's previous method can be seen in \textit{Art and
Entertainment}, which is now less confused with other categories, in
\textit{Society}, which is now less confused with
\textit{Family\_and\_parenting}, and in \textit{Health\_and\_fitness},
previously more confused with a lot of other categories containing similar
contents like \textit{Society, Sport and Food\_and\_drink}.
In computational terms there are no relevant differences, since the addition
of a 300-sized Doc2Vec vector does not change the order of magnitude of the
feature vectors to be used for the training and classification steps.

\subsection{Examples}
Last but not least, practical classification examples are reported by showing
two different Wikipedia documents. In the first example, a document that talks
about a specific mineral called ``Bukovskyite'', and was labeled in
\textit{Corpus\_B} as \textit{Iron\_and\_steel\_industry}, has been classified
correctly as \textit{business\_and\_industrial-/iron\_and\_steel\_industry/}, as
well as \textit{science/geology/} that is correct for obvious reasons.
In the second example a document talking about food-related problems
has been classified. As shown, the classifier returned categories including the
correct label \textit{food\_and\_drink/healthy\_eating/} as the second choice,
which may be considered a good result, but also contains a more relevant
category for such a document like 
\textit{health\_and\_fitness/addiction/}, which constitutes a surprising
achievement.

\subsubsection{Iron\_and\_steel\_industry Wikipedia document.} 
\textit{Bukovskyite (also known as ``clay of Kutn\'a Hora'') is an iron arsenate
sulfate mineral which forms nodules with a reniform (kidney-shaped) surface.
Under a microscope, these nodules appear as a collection of minute needles
similar to gypsum. Some can be seen with the naked eye and occur inside the
nodules. Bukovskyite was first described from pit heaps from the Middle Ages,
where sulfate ores had been mined at Kank, north of Kutn\'a Hora in Bohemia,
Czech Republic, and other old deposits in the vicinity. Only recently defined
and acknowledged, it was approved by the IMA in 1969. Bukovskyite was collected
a long time ago from the overgrown pit heaps by the inhabitants of Kutn\'a Hora.
It was used for poisoning field mice and other field vermin. This poisonous
clay, known also by the place name as ``clay of Kutn\'a Hora'', was widely known and
it was considered to be arsenic (arsenic trioxide).}
\subsubsection{Classification results}
\begin{itemize}
\item \textbf{Label} = \textit{business\_and\_industrial/iron\_and\_steel\_industry/} \textbf{Score} = 0.68
\item \textbf{Label} = \textit{science/geology/}  \textbf{Score} =  0.53
\end{itemize}

\subsubsection{Healthy eating Wikipedia document.} 
\textit{Overeaters Anonymous (OA) is a twelve-step program for people with
problems related to food including, but not limited to, compulsive overeaters,
those with binge eating disorder, bulimics and anorexics. Anyone with a
problematic relationship with food is welcomed, as OA's Third Tradition states
that the only requirement for memberships is a desire to stop eating
compulsively. OA was founded by Rozanne S. and two other women in January 1960.
The organization\'s headquarters, or World Service Office, is located in Rio
Rancho, New Mexico. Overeaters Anonymous estimates its membership at over 60,000
people in about 6,500 groups meeting in over 75 countries. OA has developed its
own literature specifically for those who eat compulsively but also uses the
Alcoholics Anonymous books Alcoholics Anonymous and Twelve Steps and Twelve
Traditions. The First Step of OA begins with the admission of powerlessness over
food; the next eleven steps are intended to bring members physical, emotional,
and spiritual healing.}
\subsubsection{Classification results}
\begin{itemize}
\item \textbf{Label} = \textit{health\_and\_fitness/addiction/} \textbf{Score}
= 0.64
\item \textbf{Label} = \textit{food\_and\_drink/healthy\_eating/} \textbf{Score} =  0.38
\item \textbf{Label} = \textit{food\_and\_drink/gastronomy/} \textbf{Score} = 0.26
\end{itemize}

\subsection{Technical configuration for the experimentation}
The hardware configuration employed for the reported experimentation includes 
the following systems: one Intel i7-6700HQ CPU with 16 GB DDR3 RAM, one Intel
i7-7700 CPU with 32GB DDR3 RAM and Sandisk Ultra SSD, and one Intel i7-8550U CPU
with 32 GB DDR4 RAM and Samsung Pro SSD. The classifier used in NETHIC has been
written in Python and exploits the {\it scikit-learn}, {\it CountVectorizer} and
{\it Multi-layer Perceptron (MLP)} libraries to create the feature vectors and
the artificial neural networks themselves. For the Doc2Vec
pre-trained model, the {\it Gensim} library has been used. Persistence and
loading of the networks is done via the {\it pickle} library.

\section{Discussion and conclusion}
\label{section:Conclusion}
This work reported and extended the discussion on NETHIC, a software tool
implementing a classification method for textual documents relying upon
hierarchical taxonomies, artificial neural networks and a document embedding
mechanism.

The earlier research discussed in~\cite{NETHIC-ICEIS2019} proved the
combination of artificial neural networks and hierarchical taxonomies to be
effective for tackling the classification problem, displaying an overall solid
performance together with relevant characteristics of scalability and
modularity.

With respect to the initial version of NETHIC, the
current tool now takes advantage of a state-of-art Natural Language Processing
technique like Doc2Vec, and the results achieved with the introduction of such
an embedding technique, in combination with the earlier used bag-of-words,
have demonstrated that with a slight increase in the
dimensional space it is possible to obtain better results in the classification
of documents and texts.

In this regard, the experimentation reported in this work showed that the
improvements obtained with respect of NETHIC's original method via the
combination of the BOW and Doc2Vec embedding mechanisms encourages their
combined usage so that more information can be considered by NETHIC's neural
networks in order for it to understand and choose the correct categories for
classification. Taken individually, the BOW mechanism proved to be sufficiently
solid (as seen in~\cite{NETHIC-ICEIS2019}), whereas it may not be advisable to
use Doc2Vec by itself, probably because semantically-close categories, like the
leaves of a given intermediate category, are difficult to be told apart without
considering the words used.

Future work may explore the possibilities of integrating and/or extending other
state-of-the art methods like BERT~\cite{BERT-2018} that are currently heading
towards ever newer and future-envisioning frontiers.

\bibliographystyle{splncs04}
\bibliography{biblio}

\end{document}